%% file: main.tex
\documentclass[runningheads]{llncs}
\usepackage[T1]{fontenc}
\usepackage{graphicx}
\usepackage{amsmath}
\usepackage{amssymb}
 \usepackage{relsize}
\usepackage{xcolor}
\usepackage{booktabs}
\usepackage{multirow}
\usepackage{hyperref}
\usepackage{svg}
\usepackage{dsfont}

\begin{document}

\title{Graph-level representations using ensemble-based readout functions}
\titlerunning{Ensemble-based readout functions}

\author{
Jakub Binkowski\inst{1}\orcidID{0000-0001-7386-5150} \and
Albert Sawczyn\inst{1}\orcidID{0000-0003-3142-4028} \and
Denis Janiak\inst{1}\orcidID{0000-0003-1859-9093} \and
Piotr Bielak\inst{1}\orcidID{0000-0002-1487-2569} \and
Tomasz Kajdanowicz\inst{1}\orcidID{0000-0002-8417-1012}
}

\authorrunning{J. Binkowski et al.}

\institute{
Department of Artificial Intelligence,\\
Wroclaw University of Science and Technology,\\
Wrocław, Poland\\
\email{jakub.binkowski@pwr.edu.pl}
}

\maketitle

\begin{abstract}
Graph machine learning models have been successfully deployed in various application areas. One of the most prominent types of models -- Graph Neural Networks (GNNs) -- provides an elegant way of extracting expressive node-level representation vectors, which can be used to solve node-related problems, such as classifying users in a social network. However, many tasks require representations at the level of the whole graph, e.g., molecular applications. In order to convert node-level representations into a graph-level vector, a so-called \textit{readout} function must be applied. In this work, we study existing readout methods, including simple non-trainable ones, as well as complex, parametrized models. We introduce a concept of ensemble-based readout functions that combine either representations or predictions. Our experiments show that such ensembles allow for better performance than simple single readouts or similar performance as the complex, parametrized ones, but at a fraction of the model complexity.

\keywords{Graph Neural Networks  \and Readout \and Pooling \and Graph Classification \and Graph Regression \and Machine Learning.}
\end{abstract}

\section{Introduction}
In recent years, machine learning has seen dramatic progress in areas where data is more complex, irregular, or structured. The abundance of methods treating objects of various structured domains eventually led to the movement of Geometric Deep Learning \cite{bronstein_geometric_2021}, which tries to characterize neural network architectures through the lens of data geometry and symmetries. In particular, research advancements in graph neural networks resulted in multiple novel architectures being able to address symmetries present in graph objects. One of the most important milestones was the development of Graph Neural Networks (GNNs), which rely on the convolution operation adapted to the irregular graph structure. GNNs can learn node representations by transforming initial input features throughout several message-passing layers. The final node representations might be further utilized in a task of interest and are often learned end-to-end using a task-related objective.

In the realm of Graph Machine Learning, one can distinguish several types of tasks. Among them, the most prevalent are link prediction, node-level prediction (classification or regression) and graph-level prediction (classification or regression). It turns out that for node-level tasks, using node representations obtained from message-passing GNNs is straightforward as those representations could be immediately passed to the layer producing final predictions per each node. However, in the case of graph-level tasks, before the classification stage, one has to obtain a global graph-level representation vector, which summarizes the entire object. The common practice is to aggregate node representations from the last or all GNN layers through a so-called \textbf{readout function} \cite{gilmer_neural_2017}. Since graph-level representations must be invariant to node permutations, a readout function has to aggregate node embeddings regardless of their ordering, i.e., for each possible node permutation, it must return the same value. This enforces a significant constraint and poses a real challenge. Since graph-level tasks are of great importance, e.g., in fields like chemistry or biology, and their performance strongly depends on the readout function, this work focuses on the evaluation of various approaches to perform readout. Additionally, we include proposed ensemble methods and discuss obtained results to give an intuition for choosing the proper readout for a problem of interest.

In the simplest scenario, one might leverage permutation-invariant functions, such as: \texttt{sum}, \texttt{mean}, or \texttt{max}. However, as shown by Xu et al. \cite{xu_how_2019}, these approaches might be suboptimal and lead to significant data loss. Along with the advancements in various aggregation schemes for GNN message-passing, many works attempted to refine the readout function for graph-level tasks. Recently, Buterez et al. \cite{buterez_graph_2022} performed a large-scale evaluation of various readouts, giving several unexpected insights. In this work, we aim to perform a follow-up study of selected readouts. In particular, we introduce an ensemble approach to readouts, in which we first perform several readout functions in parallel and then aggregate results in two scenarios: (1) representations of all readouts are aggregated, and the result is passed for prediction, (2) prediction is performed over embeddings from each readout individually, and then outcomes are aggregated.

\subsubsection{Contributions} This paper is aimed at bringing the following contributions:
\begin{enumerate}
    \item We introduce ensemble-based readout models at representation and prediction levels, which help obtain rich graph representation from nodes' representations and outperform non-invariant SOTA results obtained with MLP and GRU models.
    \item We perform a comparative study of readouts on datasets with varying sizes, characteristics, and target tasks.
    \item We analyse the obtained results concerning the computational burden of considered readout functions, giving guidelines for readout function selection.
\end{enumerate}

\section{Related work}

\subsection{Graph Neural Networks}
The early approaches, which enabled to perform tasks, like graph classification, involved graph kernels \cite{kriege_subgraph_nodate}. However, due to their limitations, e.g., the necessity of manually designing combinatorial features of a graph, as well as big advancements in deep learning, Graph Neural Networks were proposed. GNNs arose as a generalization of convolution to irregular graph structures. The groundbreaking works in graph machine learning proposed to perform graph convolution in a spectral domain, leveraging eigendecomposition of the graph Laplacian \cite{bruna_spectral_2014}, or by employing polynomial spectral filters \cite{defferrard_convolutional_2017}. Gilmer et al. \cite{gilmer_neural_2017} proposed to unify GNNs through the message-passing framework, showing that previous convolution operators are its special cases. The rapid growth in the field led to many variants of Message Passing Neural Networks (MPNNs), among which the most renowned ones are GCN \cite{kipf_semi-supervised_2017}, GAT \cite{velickovic_graph_2018}, GIN \cite{xu_how_2019}. Such MPNNs were widely used in tasks like graph property prediction (classification or regression), node property prediction (node classification or regression), or link prediction.

\subsection{Graph-level prediction}
While MPNNs can transform input features into nodes' representations, predicting at a graph level requires a readout function that summarises a graph's node embeddings into a single vector, further passed to a prediction model (classifier or regressor). Such function should be invariant to node order permutations \cite{bronstein_geometric_2021}, and it is desired to be injective for the sake of expressivity \cite{xu_how_2019}.

The primary approach for readouts leveraged simple permutation-invariant functions, including \texttt{sum}, \texttt{mean}, or \texttt{max}. Xu et al. \cite{xu_how_2019} proved the properties of these functions, showing their advantages and limitations. Although the work considered these functions in the context of neighbourhood aggregation in graph convolution, the same applies to them being readouts. Also, the authors raised the issue of function injectivity and stated that in certain situations, the \texttt{sum} operator might satisfy this property.

Recently, researchers developed more advanced approaches considering permutation invariance with respect to inputs. Most of these works aimed to perform various tasks on sets, which resemble the scenario for readouts. Zaheer et al. \cite{zaheer_deep_2018} presented the DeepSets architecture as the main framework to obtain permutation invariant representations of sets. We adopted this approach in our experimental scenario, and we will discuss it in more detail in the next section. Moreover, the SetTransformer \cite{lee_set_2019} model enabled the efficient application of Transformer \cite{vaswani_attention_2017} architecture to sets, which scales to large-sized inputs.

It is also worth noting that several works proposed to leverage local pooling, which attempts to group similar nodes together iteratively. For instance, DiffPool \cite{ying_hierarchical_2019} clusters together similar nodes in several iterations, such that in each iteration number of clusters decreases up to only 1 at the last layer, which serves as a graph representation. Besides, there are other local pooling methods, like Graph Memory Networks\cite{khasahmadi_memory-based_2020}, or based on GRACLUS \cite{rhee_hybrid_2018}. However, as shown by Mesquita et al. \cite{mesquita_rethinking_2020}, there are no significant performance benefits from using local pooling, hence in this work, we do not consider them in the experiments.

Recently, Buterez et al. \cite{buterez_graph_2022} showed results of various readouts based on large-scale evaluation, stating that in some situations losing permutation-invariance and employing models, like MLP or GRU recurrent neural network \cite{cho_properties_2014}, might bring superior performance. Moreover, they showed that general adaptive (i.e. parametrized and learnable) readouts often lead to better efficacy with sufficient data. Our work also employs MLP and GRU, however, as opposed to Buterez et al., we also propose to use node-shared parameters (which, in hand with a prediction layer, forms the DeepSets architecture \cite{zaheer_deep_2018}). Moreover, we propose leveraging an ensemble approach that utilizes several readouts in a single model, similar to the aggregation scheme in PNA graph convolution \cite{corso_principal_nodate}.

\section{Graph Neural Network Readouts} \label{sec:readouts}
Graph Neural Networks operate on graph input data, transforming initial node features into rich node representations. Formally, we denote a graph as $\mathcal{G} = (\mathbf{X}, \mathcal{A})$, where $\mathbf{X} \in \mathbb{R}^{|\mathcal{V}| \times d}$ are node features, and $A$ is an adjacency matrix describing the connection between nodes, such that $a_{uv} = \mathds{1} ((u, v) \in \mathcal{E})$ for $\mathcal{E}$ is a set of edges and $u, v \in \mathcal{V}$ are nodes. In this work, we consider unweighted edges. Further, a GNN layer is a function taking a graph as input and transforming it into a latent space -- see: Eq. \eqref{eq:gnn}. Such formulation describes a generic framework of Message Passing Neural Network. First, it aggregates the neighborhood of a node through a permutation-invariant $\mathlarger{\oplus}$ function to compute a message, and then it combines the message with the node's representation through the $\Theta$ function. Both functions could be parameterized by neural networks. Particular instances of GNNs, like GCN or GIN, vary only in the $\Theta$ and $\mathlarger{\oplus}$ functions. The whole network often comprises several such layers, and their number should be adjusted to a particular dataset.

\begin{equation} \label{eq:gnn}
\mathbf{h}_{u}^{(l+1)} = \Theta^{(l)}\left(\mathbf{h}_u^{(l)}, \mathlarger{\oplus}^{(l)}\left(\{\mathbf{h}_v^{(l)}, \forall v \in \mathcal{N}(u)\}\right)\right)
\end{equation}

\subsection{Problem statement}

\begin{figure}
    \centering
    \includegraphics[width=\textwidth]{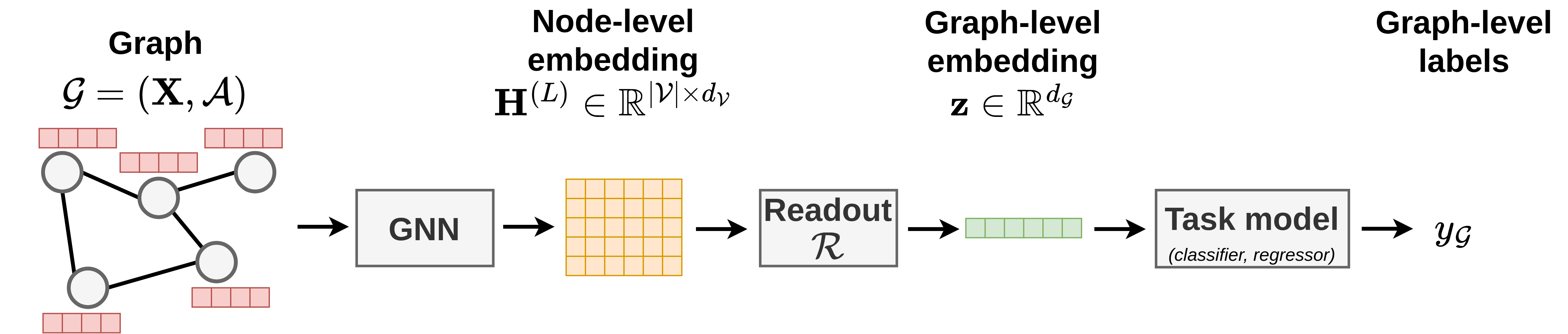}
    \caption{A readout function in the whole graph-level prediction pipeline.}
    \label{fig:readout}
\end{figure}

Formally, a readout function, denoted as $\mathcal{R}$, is an aggregation function that takes a matrix of stacked node embeddings (usually taken from the last GNN layer) at the input and returns a single vector at its output, i.e., $\mathcal{R}(\mathbf{H}^{(L)}) \rightarrow \mathbf{z}$, where $\mathbf{H}^{(L)} \in \mathbb{R}^{|\mathcal{V}| \times d_{\mathcal{V}}}$, and $\mathbf{z} \in \mathbb{R}^{d_\mathcal{G}}$, where $L$ is the number of layers, $d_\mathcal{V}$ dimension of a hidden node representation, and $d_\mathcal{G}$ dimension of graph representation. In many scenarios, the dimension of the graph representation vector is equal to the dimension of the nodes' representation, i.e., $d_\mathcal{G}=d_\mathcal{V}$. Fig. \ref{fig:readout} visually depicts the concept of the readout function.

Readout function has to be invariant to the permutation of the node order in a graph, meaning that $\mathcal{R}(\mathbf{H}^{(L)}) = \mathcal{R}(\mathbf{P}\mathbf{H}^{(L)})$, where $\mathbf{P}$ is a row-permutation matrix. As showed by Zaheer et al. \cite{zaheer_deep_2018}, permutation-invariant functions over the elements could be expressed as in Eq. \eqref{eq:deepsets}, where $\rho$ and $\phi$ could be functions parametrized by neural networks. Zaheer et al. considered such parametrization with MLP networks as the DeepSets architecture. However, it might be hard to train representations in a DeepSets setting, which was theoretically discussed by Wagstaff et al. \cite{wagstaff_limitations_2019}. Nonetheless, we include DeepSets approach in our experiments as well. Besides, there are several different approaches to performing prediction on sets, e.g., SetTransformer \cite{lee_set_2019}, or Set2Set \cite{vinyals_order_2016}.

\begin{equation}
\label{eq:deepsets}
    f(X) = \rho\left(\sum_{x \in X}\phi(x)\right)
\end{equation}

\subsection{Non-parametrized readout functions}
This work considers three basic non-parametrized functions: \texttt{sum}, \texttt{mean}, and \texttt{max}. These are the simplest realizations of readouts which satisfy the necessary invariance conditions. However, there are certain drawbacks to such aggregations. The concerns were raised by Xu et al. \cite{xu_how_2019}, who characterized each of the three aggregators with respect to their expressivity, stating that \texttt{sum} could capture the most information from the node embeddings, yet the effectiveness of such readout function might depend on the underlying data. However, as proved by Corso et al. \cite{corso_principal_nodate}, a single aggregation might not satisfy injectivity over real numbers, hence they propose to concatenate results of several aggregations at each convolutional layer. Motivated by these results, we propose an ensemble approach for building readout function. In particular, we evaluate the effectiveness of a readout function which composes \texttt{sum}, \texttt{mean}, and \texttt{max} aggregator into one vector. 

\subsection{Parametrized readout functions}
In contrast to \texttt{sum}, \texttt{mean}, and \texttt{max}, parametrized readout functions could be optimized during training. In this work, we consider three such functions: DeepSets, Virtual Node, and an ensemble approach. Also, motivated by the surprising performance of permutation-sensitive methods \cite{buterez_graph_2022}, we include MLP- and GRU-based readouts. Now, let us discuss each parametrized readout in more detail.

\subsubsection{DeepSets} We use DeepSets architecture as in Eq. \eqref{eq:deepsets}, where $\phi$ and $\rho$ are parametrized by an MLP network. This approach is a special case of the encoder-decoder architecture, in which $\phi$ is the encoder with weights shared among all nodes, and $\rho$ is the decoder transforming node representations summed over the embedding dimension. We employ two variants of DeepSets model, namely \texttt{DeepSets-Base} and \texttt{DeepSets-Large}, which differ by the number of layers in the encoder. Both versions have the same 2-layer MLP at the decoder, which predicts the final output.

\subsubsection{Virtual Node} In this approach, no explicit readout function is present in the architecture. Instead, the graph adjacency matrix is altered such that it includes one extra node connected with all other nodes in the graph \cite{gilmer_neural_2017}.  Then, the graph-level representation is the embedding of this single node trained along with the remaining node embeddings (virtual node features are initialized with zeros). The aggregation function of the graph convolution might be seen as an implicit readout, and due to its permutation-invariance, this approach is also insensitive to the modification in the node ordering.

\subsubsection{Dense and GRU} In their recent work, Buterez et al. \cite{buterez_graph_2022} empirically showed that on many datasets loosening permutation-invariance and leveraging MLP (Dense) or GRU architectures might lead to outperformance over other readouts, including the learnable ones. However, they argue that this approach is particularly suitable for data with consistent node ordering like canonical-SMILES \cite{oboyle_towards_2012} in molecular datasets. 

\subsection{Ensemble readouts} In this work, we propose to evaluate ensemble approaches to readout functions that combine results from many readouts. As described by Xu et al. \cite{xu_how_2019}, each aggregation from \texttt{sum}, \texttt{mean}, \texttt{max} has specific desirable properties, hence using three of them might bring a performance improvement. We propose to use these three readout functions and consider three approaches to aggregate representation computed with each of them. 

\begin{enumerate}
    \item First, we propose to simply concatenate the results of each aggregation, which are then forwarded to a target predictor. We denote that approach as \texttt{ConcatR}, where $||$ denotes the concatenation operator:

\begin{equation}
    \mathbf{z} = \mathcal{R}_1(\mathbf{H}^{(L)})\; || \;\mathcal{R}_2(\mathbf{H}^{(L)}) \;||\; \ldots \; ||\; \mathcal{R}_N(\mathbf{H}^{(L)}),
\end{equation}
    
    \item In the second scenario, we build the final graph-level representation as a weighted sum of each readout's output. This approach is formally expressed by Eq. \eqref{eq:ensemble_weighted}, where $\{\mathcal{R}_1, \ldots, \mathcal{R}_N\}$ is a set of $N$ readout functions, $(\mathbf{W^{(p)}_r}, \mathbf{b^{(p)}_r})$ are weights and biases of a linear projection layer specific to the readout $r$, and $(w^{(c)}_r, b^{(c)}_r)$ are weights and biases of a combination layer. Note that representations from each readout might result in different scales, so we experiment with turning the projection layer on and off. 

\begin{equation} \label{eq:ensemble_weighted}
    \mathbf{z} = \sum_{r \in \{\mathcal{R}_1, \ldots, \mathcal{R}_N\}}w^{(c)}_r  \left(\mathbf{W^{(p)}_r}\mathbf{z}_r + \mathbf{b^{(p)}_r}\right) + b^{(c)}_r
\end{equation}

    We propose two variants of such representation-level ensembles:
    \begin{itemize}
        \item \texttt{WMeanR} (\textit{Weighted Mean of Readouts}) -- the representation from each readout is multiplied by a learnable weight and summed together (the projection layer is replaced with an identity map: $\mathbf{W^{(p)}_r} = \mathds{I}$, $\mathbf{b}^{(p)}_r = \mathbf{0}$),
        \item \texttt{WMeanR+Proj} (\textit{Weighted Mean of Readouts with Projection layer}) -- the representation from each readout is first transformed by the projection layer, and then weighted and summed together (as in the previous model).
    \end{itemize}

    \item The last scenario involves the aggregation of results from the three readouts at the prediction level, as shown in Eq. \eqref{eq:ensemble_weighted_decision}. The symbols are the same as in Eq. \eqref{eq:ensemble_weighted}, and $\psi(\cdot)$ denotes a predictor model. Again, we include models with and without a projection layer in our experiments.

\begin{equation} \label{eq:ensemble_weighted_decision}
    \mathbf{\hat{y}} = \sum_{r \in \{\mathcal{R}_1, \ldots, \mathcal{R}_N\}}\left(w^{(c)}_r  \psi(\mathbf{W^{(p)}_r} \mathbf{z}_r + \mathbf{b^{(p)}_r}) + b^{(c)}_r\right)
\end{equation}

    \begin{itemize}
        \item \texttt{MeanPred} (\textit{Mean of readouts' Predictions}) -- predictions are computed for each readout separately and then averaged over all readouts (the projection layer is replaced with an identity map: $\mathbf{W^{(p)}_r} = \mathds{I}$, $\mathbf{b}^{(p)}_r = \mathbf{0}$; we also used fixed values for the combination parameters: $w^{(c)}_r = 1, b^{(c)}_r = 0$),
        \item \texttt{WMeanPred} (\textit{Weighted Mean of readouts' Predictions}) -- predictions are computed for each readout separately, then aggregated by learnable weights (the projection layer is replaced with an identity map: $\mathbf{W^{(p)}_r} = \mathds{I}$, $\mathbf{b}^{(p)}_r = \mathbf{0}$),
        \item \texttt{WMeanPred+Proj} (\textit{Weighted Mean of readouts' Predictions with Projection layer}) -- projection layer is applied to representations from each readout, then predictions are computed and aggregated by learnable weights  
    \end{itemize}

\end{enumerate}

Altogether, we introduce \textbf{6 different ensemble-based variants of readout functions}.

\section{Experiments}
Here, we describe experiments and obtained results in detail. First, we elaborate on the datasets and splits, further, we specify the experimental setting, and finally, we discuss the results.

\subsection{Datasets}
In our experiments, we utilized four datasets, which are MUTAG \cite{kriege_subgraph_nodate}, ENZYMES \cite{borgwardt_protein_2005}, ZINC \cite{bresson_two-step_2019} (12K subset of the dataset, as proposed in \cite{dwivedi_benchmarking_2022}), and REDDIT-MULTI-12K \cite{10.1145/2783258.2783417}. Datasets were selected to cover various tasks, domains, and graph sizes. Table \ref{tab:datasets} summarizes datasets statistics. For each dataset from the TUD repository (MUTAG, ENZYMES, REDDIT-MULTI-12K), we performed random splits to \textit{train}/\textit{val}/\textit{test} subsets in proportions 80\%/10\%/10\%. In the case of the ZINC dataset, we leverage pre-defined random splits without any further modifications. In the case of binary classification on the MUTAG dataset, we measure performance with the $F1$ score, for multi-class classification on ENZYMES and REDDIT-MULTI-12K datasets, we leveraged the macro-averaged $F1$ score, and for ZINC the $R^2$ metric was used.

\begin{table}[htb]
    \centering
    \caption{Tasks and statistics for each of four datasets used in the experiments.}
    \label{tab:datasets}
    \begin{tabular}{llrrr}
        \toprule
         \textbf{Dataset} & \textbf{Task} & \textbf{\# graphs} & \textbf{Avg. nodes} & \textbf{Avg. edges}  \\ 
         \midrule
         ENZYMES & multi-class & 600 & 32.9 & 125.4 \\
         MUTAG & binary & 188 & 18.0 & 39.9 \\
         ZINC & regression & 12 000 & 23.2 & 49.8 \\
         REDDIT-MULTI-12K & multi-class & 11 929 & 391.0 & 913.3 \\
         \bottomrule
    \end{tabular}

\end{table}

\subsection{Evaluation protocol}
We evaluated three different and most prevalent graph convolutions, i.e., GCN, GAT and GIN, for each readout described in section \ref{sec:readouts}. We set the hyperparameters of each model based on the ones found in \cite{dwivedi_benchmarking_2022} with slight modifications when necessary. All hyperparameters specific to each dataset are presented in Table \ref {tab:hparams}. For \texttt{Dense} and \texttt{GRU} readout, we adopted the architectures from Buterez et al. \cite{buterez_graph_2022}. For \texttt{DeepSets-Base}, we used MLP with two layers of size 128 followed by batch normalization \cite{ioffe_batch_2015} and ReLU activation, and the last layer additionally contains dropout \cite{srivastava_dropout_2014} with probability set to 0.4. For \texttt{DeepSets-Large}, we leveraged 6 such layers, followed by dropout with probability set to 0.4. The two considered \texttt{DeepSets} architectures were arranged to resemble \texttt{Dense} approach. It is worth noting that the number of trainable parameters of \texttt{Dense} method is relatively large, even when compared to \texttt{DeepSets-Large}. For the final predictor, i.e., classifier or regressor, the resultant representation of the readout function is always passed to an MLP with one hidden layer of size 128 (three layers including input and output).

Based on the well-established protocol schemes, we apply five times repeated bootstrap evaluation of each combination of graph convolution and readout. The models for each dataset were trained for a minimum of 10 epochs with early stopping on validation set loss and the patience set to 25 epochs. We leverage dynamic learning rate starting from a value of $1 \times 10^{-3}$, which was multiplied by 0.5 on each plateau with patience set to 10 epochs (minimum learning rate was set to $1 \times 10^{-6}$). Models were optimized with AdamW optimizer \cite{loshchilov_decoupled_2019}, following default PyTorch parameters (except for learning rate). In the case of classification, we used cross-entropy loss, and for regression - mean-squared error. Experiments were implemented with, \texttt{PyTorch} \cite{paszke_pytorch_2019}, \texttt{torch-geometric} \cite{fey_fast_2019} and \texttt{pytorch-lightning} \cite{falcon_pytorch_2019} libraries. To ensure reproducibility, we provide the full implementation of our experiments, containing entire hyperparameters configurations, in our public git repository \url{https://github.com/graphml-lab-pwr/ensemble-readouts}.

\begin{table}[ht]
    \centering
    \caption{Hyperparameters specification for each dataset, where $L$ is number of layers (last layer index), $d_{\mathcal{V}}$ is the hidden graph representation dimension, and $d_{\mathcal{G}}$ is the graph representation dimension.}
    \label{tab:hparams}
    \begin{tabular}{lrrr}
    \toprule
        \textbf{Dataset} & $L$ & $d_{\mathcal{V}}$ & $d_{\mathcal{G}}$  \\
        \midrule
        ENZYMES & 3 & 128 & 128 \\
        MUTAG & 3 & 64 & 128 \\
        ZINC & 3 & 128 & 128 \\
        REDDIT-MULTI-12K & 3 & 128 & 128 \\
    \bottomrule
    \end{tabular}
\end{table}

\subsection{Results}
Following the experimental setup described in the previous section, we obtain results presented in Table \ref{tab:results}, which shows the achieved efficacy for each dataset, split by graph convolution and readout (for all metrics \textit{greater=better}). We observe that in most cases the best results are obtained either by \textit{parametrized} or \textit{ensemble} methods. Among all parametrized readouts, non-invariant methods turned out to perform best only on MUTAG, which is the smallest dataset used in our experiments (containing only 188 graphs overall, and only \textit{19 graphs} in the test set). We hypothesize that such results might be caused by overfitting due to a relatively large number of trainable parameters in these models and a fixed node ordering. Therefore, we consider such methods as the worst choice for the readout function. On the other hand, the \texttt{DeepSets} architecture outperformed other models for ENZYMES using all types of GNNs, and for ZINC using GCN and GAT. The usage of a \texttt{Virtual Node} did not lead to any improvement.

Furthermore, the results showed that \textbf{ensemble} approaches outperformed other readouts for REDDIT-MULTI-12K dataset using GCN and GIN layers, as well as using the GIN layer on the ZINC dataset. It is worth noting that when using GIN convolution, the best ensemble approach is consistently the best overall or nearly the best. We did not observe such consistent results for GCN and GAT, yet the \textbf{ensemble} approaches provide better or comparative metric values. 

Whether to use ensembles over representations or predictions depends on the dataset and graph convolution. Each ensemble enables an additional interpretability property, since the trained weights might be further analyzed to assess the contribution of each readout method. 

Regarding the projection layer considered in these models, we observe that projection layers in ensembles do not guarantee better performance and could be omitted for the sake of decreasing model size. Nonetheless, when using a combination of various readouts from non-aligned embedding spaces, one should always verify the necessity of additional projection layers. We also tested a combination of DeepSets with \textit{non-parameterized} methods in an \textit{ensemble} setting, but we did not observe an increase in performance, hence we omitted results for brevity.

\begin{table}[thb]
    \centering
    \caption{Results of the conducted experiments. Each cell contains mean and standard deviation in percent of the metric over 5 runs. Metric $F1$ score was used for ENZYMES, MUTAG, REDDIT-MULTI-12K, and $R^2$ for ZINC (for all metrics $greater=better$). Best results are presented in \textbf{bold}, best within a class (\textit{non-parametrized}, \textit{parametrized}, \textit{ensemble}) are presented with \underline{underline}. \textit{NON-PAR}, \textit{PAR} and \textit{ENS} denote non-parametrized, parametrized and ensemble-based readout functions, respectively.}
    \label{tab:results}
    \resizebox{\textwidth}{!}{\input{tables/results_ENZYMES_MUTAG.tex}} 
    \resizebox{\textwidth}{!}{\input{tables/results_REDDIT_ZINC.tex}}
\end{table}

\begin{figure}
    \centering
    \includegraphics[width=\textwidth]{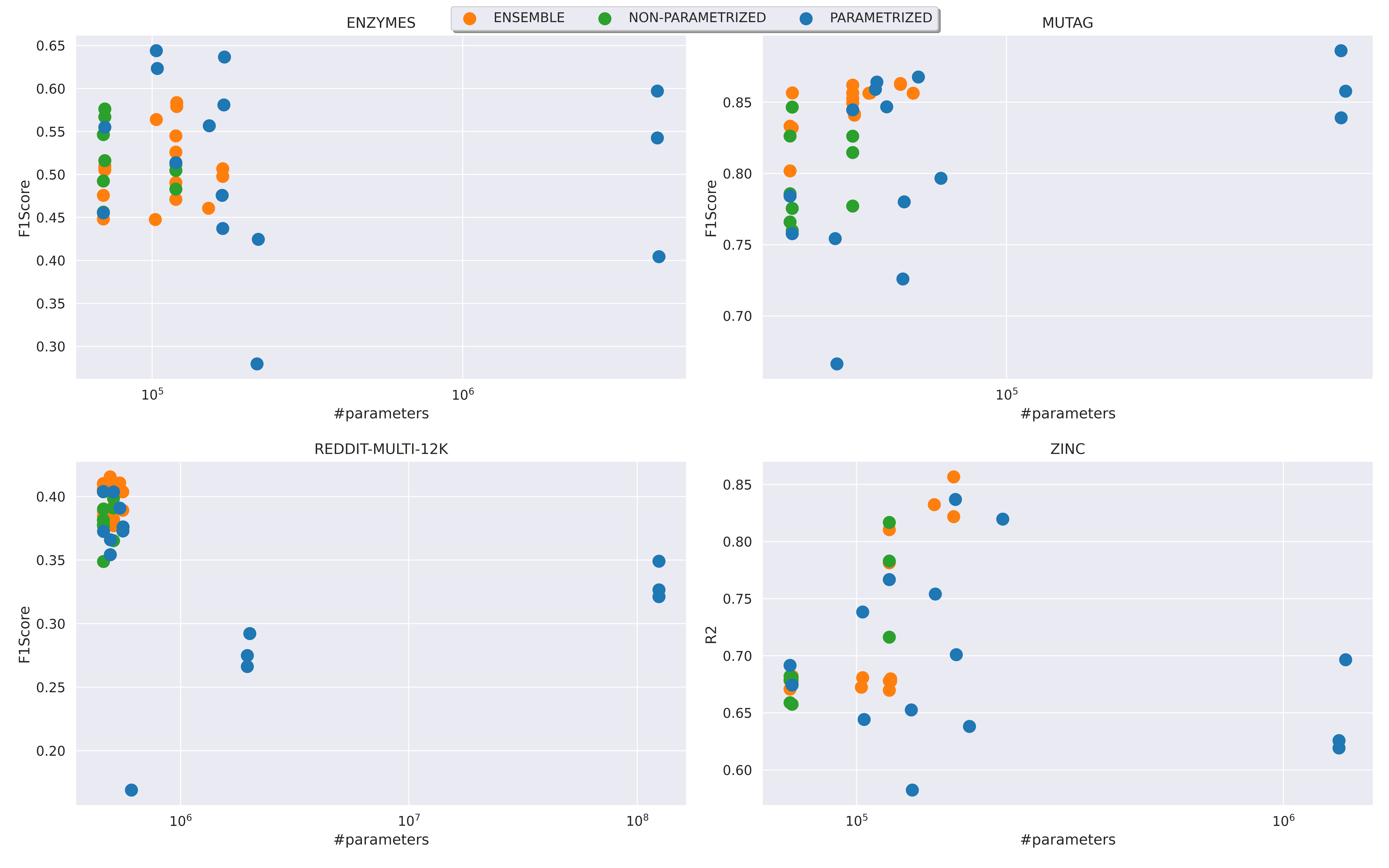}
    \caption{Comparison of the number of parameters with achieved efficacy for the four considered datasets.}
    \label{fig:params_vs_efficacy}
\end{figure}

In order to discuss the results from the perspective of model size, we plotted the efficacy achieved by models on each dataset against the number of trainable parameters, presented in Fig. \ref{fig:params_vs_efficacy}. Without accounting for a specific GNN type, the plots reveal that we can often benefit from \textbf{ensemble} or \textbf{parametrized} approaches which are comparable in size and efficacy. The exception is the dense model represented by the rightmost outlier points, which in connection with a lack of permutation-invariance position it as a poor choice for a readout function.

\section{Conclusion}
In this work, we evaluated various neural network readouts on several datasets covering different domains, cardinalities, graph properties, and tasks. We introduced \textbf{ensemble}-based approach that combines basic \textit{non-parametrized} readout functions in several ways and evaluated them against other \textit{parametrized} methods known in the literature, including DeepSets, Virtual Node, and non-invariant MLP and GRU. We showed that DeepSets and \textbf{ensemble} approaches outperform other methods in the majority of experiments and hence should be a first choice when designing models for graph-level predictions. We also do not recommend using non-invariant readouts, as they introduce larger models, might lead to overfitting, and not necessarily lead to better generalization. For  future work, we suggest investigating probabilistic modelling of node representations, including Bayesian approach, to prevent information loss, which might happen while using deterministic readouts presented in this work.

\subsubsection{Acknowledgments} 

This work was financed by (1) the Polish Ministry of Education and Science, CLARIN-PL; (2) the European Regional Development Fund as a part of the 2014–2020 Smart Growth Operational Programme, CLARIN — Common Language Resources and Technology Infrastructure, project no. POIR.04.02.00-00C002/19; (3) the statutory funds of the Department of Artificial Intelligence, Wroclaw University of Science and Technology, Poland; (4) Horizon Europe Framework Programme MSCA Staff Exchanges grant no. 101086321 (OMINO).

\bibliographystyle{splncs04}
\bibliography{main}

\end{document}

%% file: tables/results_ENZYMES_MUTAG.tex
\begin{tabular}{llllllll}
\toprule
& & \multicolumn{3}{c}{\textbf{ENZYMES} (F1 $\uparrow$)} & \multicolumn{3}{c}{\textbf{MUTAG} (F1 $\uparrow$)} \\
& & GAT & GCN & GIN & GAT & GCN & GIN\\
\midrule

\multirow[c]{3}{*}{NON-PAR} & \texttt{max} & $ 56.68 \pm  7.43$ & \underline{$ 54.63 \pm  4.86$} & $ 48.29 \pm  4.33$ & $ 77.55 \pm  7.40$ & $ 78.57 \pm  6.94$ & \underline{$ 82.61 \pm  9.39$} \\
& \texttt{mean} & \underline{$ 57.60 \pm  7.31$} & $ 45.62 \pm  6.74$ & $ 50.46 \pm  2.77$ & $ 75.99 \pm  9.52$ & $ 76.60 \pm  8.59$ & $ 77.70 \pm  8.66$ \\
& \texttt{sum} & $ 51.61 \pm  12.71$ & $ 49.23 \pm  8.08$ & \underline{$ 51.19 \pm  9.65$} & \underline{$ 84.65 \pm  6.87$} & \underline{$ 82.62 \pm  6.28$} & $ 81.47 \pm  8.83$\\
\midrule

\multirow[c]{5}{*}{PAR} & \texttt{DeepSets-Base} & $ 62.32 \pm  10.09$ & $\mathbf{\underline{64.39 \pm  6.26}}$ & $\mathbf{\underline{55.66 \pm  9.77}}$ & $ 66.63 \pm  28.74$ & $ 75.42 \pm  13.24$ & $ 84.68 \pm  6.86$ \\
& \texttt{DeepSets-Large} & $\mathbf{\underline{63.65 \pm  7.01}}$ & $ 58.08 \pm  11.49$ & $ 42.44 \pm  11.31$ & $ 78.00 \pm  3.93$ & $ 72.59 \pm  25.46$ & $ 79.66 \pm  5.97$ \\
& \texttt{Dense} & $ 59.70 \pm  7.58$ & $ 54.24 \pm  5.32$ & $ 40.43 \pm  4.15$ & $ 83.91 \pm  3.89$ & $\mathbf{\underline{88.61 \pm  8.58}}$ & $ 85.77 \pm  3.26$ \\
& \texttt{GRU} & $ 43.72 \pm  7.39$ & $ 47.56 \pm  3.75$ & $ 27.96 \pm  6.42$ & $\mathbf{\underline{86.41 \pm  5.74}}$ & $ 85.90 \pm  5.30$ & $\mathbf{\underline{86.76 \pm  4.02}}$ \\
& \texttt{Virtual Node} & $ 55.49 \pm  10.04$ & $ 45.54 \pm  3.99$ & $ 51.37 \pm  8.42$ & $ 75.77 \pm  8.74$ & $ 78.41 \pm  6.73$ & $ 84.46 \pm  7.14$ \\
\midrule

\multirow[c]{6}{*}{ENS} & \texttt{ConcatR} & $ 56.38 \pm  8.12$ & $ 44.76 \pm  10.68$ & $ 46.06 \pm  9.24$ & \underline{$ 85.65 \pm  7.03$} & $ 85.63 \pm  8.36$ & $ 85.63 \pm  8.36$ \\
& \texttt{WMeanR} & $ 50.53 \pm  8.03$ & $ 44.82 \pm  7.43$ & $ 49.04 \pm  3.65$ & \underline{$ 85.65 \pm  7.03$} & $ 83.32 \pm  8.73$ & $ 85.26 \pm  8.27$ \\
& \texttt{WMeanR+Proj} & $ 57.90 \pm  5.75$ & $ 47.09 \pm  11.09$ & $ 49.77 \pm  9.11$ & $ 84.09 \pm  6.93$ & \underline{$ 85.65 \pm  7.03$} & \underline{$ 86.31 \pm  6.08$} \\
\cmidrule{2-8}
& \texttt{MeanPred} & $ 50.95 \pm  9.59$ & $ 47.56 \pm  11.81$ & \underline{$ 54.49 \pm  5.51$} & $ 83.21 \pm  5.61$ & $ 80.18 \pm  9.37$ & $ 86.20 \pm  8.39$ \\
& \texttt{WMeanPred} & $ 51.89 \pm  10.17$ & $ 50.06 \pm  9.43$ & $ 48.52 \pm  8.90$ & $ 82.10 \pm  7.16$ & $ 79.62 \pm  10.43$ & $ 84.58 \pm  8.13$ \\
& \texttt{WMeanPred+Proj} & \underline{$ 58.35 \pm  9.43$} & \underline{$ 52.59 \pm  7.55$} & $ 50.67 \pm  10.82$ & $ 84.18 \pm  8.35$ & $ 84.95 \pm  8.26$ & $ 86.25 \pm  6.95$ \\
 \bottomrule
\end{tabular}

%% file: tables/results_REDDIT_ZINC.tex
\begin{tabular}{llllllll}
\toprule
& & \multicolumn{3}{c}{\textbf{REDDIT-MULTI-12K} (F1 $\uparrow$)} & \multicolumn{3}{c}{\textbf{ZINC} ($R^2 \;\uparrow$)} \\
& & GAT & GCN & GIN & GAT & GCN & GIN \\
\midrule

\multirow[c]{3}{*}{NON-PAR} & \texttt{max} & $ 37.66 \pm  1.56$ & $ 38.15 \pm  2.95$ & $ 39.09 \pm  2.12$ & $ 65.74 \pm  0.76$ & $ 65.88 \pm  1.43$ & $ 71.62 \pm  3.41$ \\
& \texttt{mean} & $ 34.89 \pm  2.73$ & $ 37.77 \pm  1.77$ & $ 36.53 \pm  0.55$ & \underline{$ 68.10 \pm  0.57$} & $ 67.84 \pm  0.79$ & $ 78.29 \pm  2.85$ \\
& \texttt{sum} & $\mathbf{\underline{39.02 \pm  1.16}}$ & \underline{$ 40.38 \pm  0.97$} & \underline{$ 39.86 \pm  0.92$} & $ 67.88 \pm  0.89$ & \underline{$ 68.21 \pm  0.26$} & \underline{$ 81.68 \pm  1.65$} \\
\midrule

\multirow[c]{5}{*}{PAR} & \texttt{DeepSets-Base} & $ 36.59 \pm  2.35$ & $ 35.42 \pm  2.75$ & $ 39.08 \pm  1.82$ & $ 64.42 \pm  9.36$ & $ 73.82 \pm  2.39$ & $ 75.40 \pm  2.17$ \\
& \texttt{DeepSets-Large} & \underline{$ 37.61 \pm  1.71$} & $ 37.29 \pm  2.06$ & $ 16.90 \pm  10.11$ & $\mathbf{\underline{70.09 \pm  2.77}}$ & $\mathbf{\underline{83.69 \pm  3.26}}$ & \underline{$ 81.96 \pm  1.24$} \\
& \texttt{Dense} & $ 32.12 \pm  1.67$ & $ 32.65 \pm  1.10$ & $ 34.91 \pm  1.36$ & $ 61.93 \pm  1.73$ & $ 62.57 \pm  1.31$ & $ 69.65 \pm  1.39$ \\
& \texttt{GRU} & $ 27.49 \pm  1.38$ & $ 26.62 \pm  1.49$ & $ 29.21 \pm  2.40$ & $ 58.23 \pm  8.00$ & $ 65.26 \pm  1.20$ & $ 63.81 \pm  2.91$ \\
& \texttt{Virtual Node} & $ 37.26 \pm  1.60$ & \underline{$ 40.39 \pm  0.44$} & \underline{$ 40.37 \pm  1.21$} & $ 67.43 \pm  1.44$ & $ 69.16 \pm  0.31$ & $ 76.67 \pm  3.01$ \\
\midrule

\multirow[c]{6}{*}{ENS} & \texttt{ConcatR} & $ 38.62 \pm  1.90$ & $\mathbf{\underline{41.54 \pm  0.75}}$ & $\mathbf{\underline{41.06 \pm  1.05}}$ & $ 68.08 \pm  0.56$ & $ 67.24 \pm  1.33$ & $ 83.23 \pm  2.56$ \\
& \texttt{WMeanR} & \underline{$ 38.83 \pm  0.94$} & $ 41.02 \pm  0.75$ & $ 39.22 \pm  2.58$ & $ 67.66 \pm  0.88$ & $ 67.08 \pm  0.82$ & $ 81.04 \pm  2.00$ \\
& \texttt{WMeanR+Proj} & $ 37.70 \pm  1.04$ & $ 40.65 \pm  1.17$ & $ 40.37 \pm  1.15$ & $ 67.97 \pm  0.63$ & $ 67.80 \pm  0.26$ & $ 82.18 \pm  2.31$ \\
\cmidrule{2-8}

& \texttt{MeanPred} & $ 38.46 \pm  1.52$ & $ 40.59 \pm  1.79$ & $ 40.20 \pm  1.25$ & \underline{$ 68.22 \pm  0.58$} & $ 67.90 \pm  0.47$ & $ 78.14 \pm  2.24$ \\
& \texttt{WMeanPred} & $ 37.56 \pm  1.30$ & $ 39.91 \pm  0.98$ & $ 38.62 \pm  2.02$ & $ 67.28 \pm  1.90$ & \underline{$ 68.22 \pm  0.37$} & $ 80.06 \pm  3.41$ \\
& \texttt{WMeanPred+Proj} & $ 38.18 \pm  1.11$ & $ 40.65 \pm  1.95$ & $ 38.92 \pm  1.34$ & $ 67.77 \pm  1.19$ & $ 66.98 \pm  2.22$ & $\mathbf{\underline{ 85.66 \pm  2.70}}$ \\
\bottomrule
\end{tabular}